\documentclass{article}

\usepackage{PRIMEarxiv}

\usepackage[utf8]{inputenc} 
\usepackage[T1]{fontenc}    
\usepackage{hyperref}       
\usepackage{url}            
\usepackage{booktabs}       
\usepackage{amsfonts}       
\usepackage{amsmath}
\usepackage{nicefrac}       
\usepackage{microtype}      
\usepackage{lipsum}
\usepackage{fancyhdr}       
\usepackage{graphicx}       
\usepackage{soul}
\usepackage{xcolor}
\usepackage{makecell}
\usepackage{longtable}
\usepackage{pgfplotstable}
\usepackage{adjustbox}
\usepgfplotslibrary{statistics}

\pgfplotsset{compat=1.18}
\title{OSUniverse: Benchmark for Multimodal GUI-navigation AI Agents
}

\author{
  Mariya Davydova\textsuperscript{*}, Daniel Jeffries\textsuperscript{*}, Patrick Barker, Arturo Márquez Flores, Sinéad Ryan \\
  \\
  Kentauros AI Inc. \\
  \texttt{research@kentauros.ai} \\
}

\begin{document}
\maketitle

\begingroup
\renewcommand\thefootnote{*}
\footnotetext{Equal contribution}
\endgroup

\begin{abstract}
In this paper, we introduce OSUniverse: a benchmark of complex, multimodal desktop-oriented tasks for advanced GUI-navigation AI agents that focuses on ease of use, extensibility, comprehensive coverage of test cases, and automated validation. We divide the tasks in increasing levels of complexity, from basic precision clicking to multistep, multiapplication tests requiring dexterity, precision, and clear thinking from the agent. In version one of the benchmark, presented here, we have calibrated the complexity of the benchmark test cases to ensure that the SOTA (State of the Art) agents (at the time of publication) do not achieve results higher than 50\%, while the average white collar worker can perform all these tasks with perfect accuracy. The benchmark can be scored manually, but we also introduce an automated validation mechanism that has an average error rate less than 2\%. Therefore, this benchmark presents solid ground for fully automated measuring of progress, capabilities and the effectiveness of GUI-navigation AI agents over the short and medium-term horizon. The source code of the benchmark is available at \href{https://github.com/agentsea/osuniverse}{https://github.com/agentsea/osuniverse}.

\end{abstract}


\pgfplotsset{
  agentplot/.style={
    xbar,
    width=0.7\textwidth,        
    height=8cm,                 
    xlabel={Score (\%)},
    symbolic y coords={8,7,6,5,4,3,2,1},
    yticklabels={
        {computer-use-preview-2025-03-11},
        {claude-3-5-sonnet-20241022(computer-use)},
        {claude-3-5-sonnet-20241022(agentdesk)},
        {qwen2.5-vl-72b-instruct},
        {gemini-2.5-pro-exp-03-25},
        {gemini-2.0-flash-001},
        {gpt-4o-2024-11-20},
        {gemini-1.5-pro-002}
    },
    ytick=data,
    yticklabel style={
        font=\small
    },
    bar width=6pt,
    nodes near coords,
    xmin=0,
    xmax=100
  }
}

\pgfplotsset{
  agentplottotal/.style={
    xbar,
    width=0.7\textwidth,        
    height=8cm,                 
    xlabel={Weighted Score (\%)},
    symbolic y coords={8,7,6,5,4,3,2,1},
    yticklabels={
        {computer-use-preview-2025-03-11},
        {claude-3-5-sonnet-20241022(computer-use)},
        {claude-3-5-sonnet-20241022(agentdesk)},
        {qwen2.5-vl-72b-instruct},
        {gemini-2.5-pro-exp-03-25},
        {gemini-2.0-flash-001},
        {gpt-4o-2024-11-20},
        {gemini-1.5-pro-002}
    },
    ytick=data,
    yticklabel style={
        font=\small
    },
    bar width=8pt,
    nodes near coords,
    xmin=0
  }
}

\section{Introduction}

Multimodal GUI-navigation agents are AI systems that perceive graphical user interfaces (GUIs) (via vision and sometimes text) and interact with a desktop using a mouse, touch, or keyboard actions to perform tasks in a similar way as a human interacts with a desktop. Purely multimodal agents are different from agents that do browser-based activities and rely on working with the underlying code, such as JavaScript DOM scraping to exact data from a web page's Document Object Model (DOM), to augment their visual reasoning.  While these code scraping agents, such as Browser Use \cite{browser_use2024}, have proven effective for many browsing tasks, their capabilities do not carry over to desktop applications because these applications rely on that code for their action accuracy.  

During the last decade, many benchmarks have been proposed to evaluate multimodal agents, both in academia and industry. These benchmarks typically measure an agent’s task success rate or completion accuracy, sometimes speed / efficiency (e.g., time or steps taken to complete tasks), and robustness or generalization to new interfaces. However, most of these benchmarks fail to capture the richness and complexity of how humans interact with computers, a complexity people take for granted because it seems so easy and intuitive to them.  Yet complex app usage and GUI navigation remains incredibly challenging for machines, involving reasoning, visual perception, and precise action execution.  GUI navigation is a classic example of \textbf{Moravec's paradox} – which essentially states that things that are easy for humans are difficult for robots, and vice versa.  Any mistakes can cause cascading errors into the task that are challenging for the agent to recover from, such as clicking out of a long wizard-based sequence and needing to start over and redo many steps, or more serious errors like adding too many items to a cart so a user is overcharged for something they don't need in a checkout sequence. 

Despite the growing popularity of these multimodal agent systems, there are very few benchmarks designed for them. In addition, these benchmarks fail to capture that nuance of human-computer interactions or reflect real-world tasks, such as moving fluidly between applications, synthesizing information from one application and converting or compressing it for use in other applications, or copying selective pieces of information over to a new application. Even office workers with minimal experience and training do incredibly complex tasks with ease, and there is a need to capture these kinds of tasks better with a benchmark. 

We highlight three well-known benchmarks in the list below. The first two are web-based, while the third, OSWorld, is the only desktop-oriented benchmark. We also highlight some of the limitations of each of them and why a new benchmark is needed to better capture the skills of human office workers. 

\textbf{\textit{WebShop (2022)}} \cite{yao2022webshop}: An academic benchmark from Princeton that simulates an e-commerce website with 1.18M real products and more than 12k crowd-sourced instructions.  An agent must parse a text instruction (for example, "find and buy a red dress under \$50"), then navigate through search pages, product listings, and item pages to find and purchase the correct item. WebShop tests language grounding in a realistic web setting, including query formulation and handling noisy text. Success is measured by whether the correct item is purchased (task success rate), and the benchmark also considers the number of steps (clicks/searches) as a measure of efficiency. In the initial evaluation, the best learning-based agent had only a 29\% success rate, versus 59\% for human experts on the same tasks. A simple rule-based heuristic achieved ~9.6\%, highlighting the task’s difficulty. This gap between human performance underlines that agents still struggle with complex, multi-step web tasks. 

\textbf{\textit{Mind2Web (2023)}} \cite{deng2023mind2web}: A large-scale academic dataset (NeurIPS 2023) for generalist web agents, with 2,350 tasks collected from 137 real websites across 31 domains. Tasks are open-ended (e.g., making travel reservations, using social media) and come with crowd-sourced action sequences as demonstrations. Unlike MiniWoB’s \cite{humphreys2022datadrivenapproachlearningcontrol} simulated pages, Mind2Web uses actual websites (with HTML and dynamic content), aiming to test an agent’s ability to generalize to unseen websites and tasks. Performance is evaluated by task completion rate (did the agent reach the goal) and also intermediate checkpoint accuracy in some evaluations. Because real web pages are large, solutions often involve first filtering the DOM or screenshot before feeding it to a model. Initial baselines using large language models (LLMs) showed moderate success. For example, a GPT-4 based agent achieved about 48.8\% task completion on a live subset of Mind2Web, but with only 23.1\% strict success rate when requiring every intermediate step to be correct. The authors note there is “substantial room to improve towards truly generalizable agents." Mind2Web also introduced robustness tests – e.g., evaluating on entirely new websites or domains the agent never saw in training – to gauge generalization.

\textbf{\textit{OSWorld (2024)}} \cite{xie2024osworld}: A recent academic benchmark (NeurIPS 2024) that provides a full-fledged computer environment (a real Operating System (OS)) for evaluating multimodal agents. OSWorld includes a suite of 369 diverse tasks on Ubuntu and Windows, spanning desktop application control, file operations, web browsing, and multi-app workflows. Each task is derived from real use cases (with a detailed initial state and an automated evaluation script to check success). For example, tasks include sending an email via a mail client, editing a spreadsheet, or downloading and opening a file from the Web. Evaluation is execution-based – the agent must actually carry out the steps in a virtual machine – and performance is measured by success or failure on each task (with all-or-nothing criteria defined by the script). Human testers can solve about 72.4\% of the OSWorld tasks, whereas the best AI agent (a state-of-the-art vision-language model augmented for actions) succeeded on only 12.2\% of tasks at the time of the OSWorld paper publication. At the time of publication of this article, the best reported result in OSWorld is 38.1\%. This stark gap indicates current agents struggle with complex GUI grounding and procedural knowledge in general computer use.

Since its publication, \textbf{\textit{OSWorld}} has become renowned as a very hard benchmark that proves incredibly challenging for even SOTA GUI-navigation AI agents. However, while this benchmark deserves credit for providing a challenging set of tasks and being the first to test both browsing and OS capabilities, it has several major shortcomings, which we highlight below:

\begin{itemize}
    \item It is challenging to run this benchmark; the environments are executed in a proprietary VMWare Workstation or VirtualBox virtual machine (or via their Docker support, which simply runs the KVM VM inside Docker), which makes each run quite resource intensive and challenging to set up.
    \item The benchmark is largely limited to ReACT-style agents and it is hard to implement any other kind of agent, which limits its usefulness as a SOTA agent may have a breakthrough that does not use a ReACT-style framework: the benchmark executor accepts only one agent interaction model and does not allow editing the agent architecture or even prompts; it is possible to write custom executors and inject that into the benchmark architecture; however, this possibility is not documented and therefore requires a lot of effort to figure out and implement through reverse engineering, which is not ideal.
    \item The tests require deterministic validation; as many of the tests deal with real-life dynamic environments (like websites of various companies), deterministic validation is prone to errors, thus adding noise to the benchmark scores. In addition, many useful agentic use cases that are worth testing simply cannot be tested deterministically (for example, searching for accommodation, flights, or goods).
    \item Finally, the prompts for the benchmark are vague and require a substantial amount of inductive and deductive reasoning for a machine to even interpret what to do, much less perform its task. While it is valid to test the reasoning of models, a GUI navigation benchmark should have clear, precise tasks laid out for the model and not require massive intuitive leaps to even understand the prompt. If a person were given such a prompt, they would be able to ask clarifying questions, or research steps to take to achieve their goal, both of which are not possible for the model in the benchmark, which significantly hinders its performance, resulting in much lower scores – not because a model cannot do the task in many cases, but because it cannot understand the prompt.
\end{itemize}

We designed \textbf{OSUniverse} to overcome these limitations and provide a robust framework for the next generation of GUI-navigation agents.

\section{Overview} 

In the domain of benchmarking the GUI-navigation AI Agents, there are several principal components:

\begin{itemize}
    \item The \textbf{environment} where the agent is operating: in most cases, we are talking about virtual machines (such as KVM or VMware Workstation) or containers (such as Docker) with pre-installed software, but some agents run directly on the user's machine as well.
    \item The \textbf{action space}: the kinds of action an agent can take. GUI navigation involves mouse and keyboard operations.
    \item The \textbf{observation space}: the kinds of signal an agent can receive from the environment. As we are talking about general-purpose GUI navigation, we currently rely on screenshots alone, though in the future these systems may rely on continuous video streams. However, some more specialized agents may rely on additional information, such as the DOM of the web page, as noted earlier.
    \item The \textbf{agentic architecture}: the way in which the agent is built; one of the most popular architectures today is ReACT, but there are many others.
    \item The \textbf{agentic model(s)}: as there are many possible architectures, there are also many possible machine learning models (typically visual LLMs or omni-models though other architectures like ConvNets are possible) to power those architectures as the "brain" of the agent; it is worth noting that some models work better than others for certain architectures, and also there are architectures that require multiple models at once (multi-agent consensus architectures) and there are architectures that use one model.
    \item The \textbf{agentic runtime}: the actual framework that runs the agent in the environment.
    \item The \textbf{validator}: This component is crucial for ensuring the stability and representability of the use-cases.
\end{itemize}

The majority of the benchmarks out there lock almost all of these components, allowing someone to choose only the agentic model, such as Claude 3.5 or OpenAI's o3 or Llama 3.2. In reality, though, even the agentic models are not perfectly substitutable inside one architecture, as there are nuances to strong and efficient prompting from model to model, and some models are better or worse at tool calling, memory usage or returning accurate JSON, to name a few of the challenges. Benchmarking custom architectures becomes a significant issue with this level of variability in both the architectures and the models themselves.

In OSUniverse, our aim is to support as many combinations of the components as  possible, by abstracting out communication to the desktop environment to allow the most flexibility for testing:

\begin{itemize}
    \item The default \textbf{environment} is AgentDesk: a framework that allows virtual desktops to run in Docker containers. However, one may prefer to use a different environment; in this case, using the suggested Docker image as a base is recommended.
    \item The \textbf{action space} is connected to the environment. The custom agent implementations (see below) contain converters between the agent action space and the AgentDesk action space.  
    \item The \textbf{observation space} is locked for screenshots only.
    \item The default \textbf{agentic runtime} is SurfKit \cite{surfkit2024}: we support running any SurfKit-compatible agents from the YAML config or from the Docker image. However, one may implement a custom runtime and use it, as long as the artifacts of the agent test case runs are the ones expected by the validator (see below).
    \item For the \textbf{validation}, we offer automatic validation (which uses Gemini models due to the size of their context window and their excellent reasoning capabilities) and human validation (we provide a handy Streamlit UX to go through the test case runs and mark them as successes or failures) for precision accuracy.
\end{itemize}

The agentic architecture is the most open-ended component of our benchmark. A user may run a benchmark with:

\begin{itemize}
\item One of the reference agents we provide: \begin{itemize}
\item Claude-based ReACT agent \begin{itemize}
\item Requires Anthropic API key
\end{itemize}
\item QWEN 2.5 VL ReACT agent
\begin{itemize}
\item Requires Alibaba API key \end{itemize}
\item CUA (Computer Use Agent from OpenAI) ReACT agent \begin{itemize}
\item Requires OpenAI API key
\end{itemize}
\end{itemize}
\begin{itemize}
\item A baseline ReACT agent compatible with any model that can return structured JSON output
\end{itemize}
\item Any SurfKit-compatible agent
\item Any other agent that is wrapped into the Runner interface (see below).
\end{itemize}

\section{Architecture}
\label{sec:architecture}

The architecture of OSUniverse is based on running tests in Docker containers and validating them using the Gemini 2.0 (gemini-2.0-flash-001) or 2.5 (gemini-2.5-pro-preview-03-25) model with their long context windows. The benchmark consists of several components:

\begin{itemize}
    \item \textbf{Test cases}: the test units are defined in YAML files and contain all the information required to run and validate the test: the environment container and the setup, the task, and the checks (detailed next). The test cases are also divided by categories and levels, which are described in the following.
    \item \textbf{Checks}: the validation units. Each test case has to contain at least one check; the amount of checks is unlimited, and the test is only considered a pass if ALL checks pass. The following checks are available:
    \begin{itemize}
        \item \textbf{ReturnedResultCheck}: compares the result an agent returned (as text) with the expected result. This type of check is perfect for  Q\&A-type  tests: for example, an agent has to look up some information and return it to the user.
        \item \textbf{FinalScreenshotCheck}: compares the final state screenshot with the description of the expected state. This type of check fits use cases where the final state of the screen clearly defines the success criteria: for example, an agent has to open a certain application or a file.
        \item \textbf{CommandOutputCheck}: compares the output of an arbitrary bash command with the description of the expected result. This type of check works very well when the result of the agent’s work can be represented in a structured format: for example, if an agent has to edit an ODT file, we can convert it to Markdown in a check and send this Markdown to the validator.
        \item \textbf{ExpectedFlowCheck}: compares the complete flow (all steps with images, actions, and comments) with the description of the expected flow. This type of check is required for complex scenarios that involve going through several stages: for example, an agent may have to open a terminal, do some work in it, and close the terminal.
    \end{itemize}
    \item \textbf{Runners}: the modules that take the TestCase and produce the TestCaseRun. This data structure contains all the information about the test case, plus configuration data, plus the execution trajectory (the sequence of actions and related desktop screenshots). The following runner is supported:
    \begin{itemize}
        \item \textbf{SurfKitAgentRunner} runs any SurfKit-compatible agent, either from the YAML configuration or from a Docker image. This runner creates the AgentDesk desktop in a Docker container, runs the setup commands from the test case configuration, executes the task with a given agent, and generates the trajectory from the agent actions and threads (depending on the data that the agent produces during the run).
    \end{itemize}
    \item \textbf{Validators}: the modules that take the TestCaseRun and score them. The following validator is supported:
    \begin{itemize}
        \item \textbf{COTGeminiValidator} uses Gemini as a scoring model and introduces checklist-based thinking on the check result.
    \end{itemize}
    \item \textbf{Benchmark}: the module that orchestrates the running and validation of a given set of test cases. Allows parallel execution of test cases. As each test case is completely independent from other test cases, we can parallelize the process as much as our resources permit. 
    \item \textbf{Viewer}: the Streamlit application that allows reviewing the runs, including the trajectories and scores, assigning a human score (optionally) and observing the summary information about the benchmark run.
\end{itemize}

\section{Classification of Test Cases}

We classify the tests using two dimensions: the level and the category.

The \textbf{category} is the core application that this test case addresses: browser, desktop, terminal, games, gimp, LibreOffice application, or a combination of applications, which we refer to as "multiapp." 

The \textbf{level} is the measure of the complexity of the test. To pass tests of a certain level, the agent should possess specific skills, which become increasingly complex at each level. However, all tasks remain simple for an average white-collar office worker. 

Table~\ref{tab:task_levels} presents the requirements for each level, the weight of the tasks, the maximum recommended number of steps per task execution, and the number of tests of this level present in the benchmark.

\begin{longtable}{p{2cm} p{7cm} c c c}

\toprule
\textbf{Level} & \textbf{Requirements} & \textbf{$\text{Weight}_{\text{level}}$} & \textbf{Max Steps} & \textbf{\# of Tests} \\
\midrule
\endfirsthead

\toprule
\textbf{Level (cont)} & \textbf{Requirements} & \textbf{$\text{Weight}_{\text{level}}$} & \textbf{Max steps} & \textbf{\# of Tests} \\
\midrule
\endhead

\textbf{Paper} & \begin{itemize}
    \item “See” the screen through screenshots
    \item Understand the task
    \item Return the result corresponding to the task
\end{itemize} & 0.5 & 5 & 11 \\
\hline \\

\textbf{Wood} & \begin{itemize}
  \item The current screen represents the state (no scrolling or opening popup windows required)
  \item Identify positions of clearly defined GUI elements (buttons, text inputs, labels, icons, etc) and interact with them
  \item Short quick tasks
\end{itemize} & 1 & 25 & 58 \\
\hline \\

\textbf{Bronze} &  \begin{itemize}
  \item The tasks take longer than on Wood level
  \item Positioning and interacting with calendars and grids
  \item Scrolling or opening dialog windows may be required (i.e. part of the state is hidden)
\end{itemize} & 2 & 50 & 48 \\
\hline \\

\textbf{Silver} & \begin{itemize}
  \item Tasks have several distinct subtasks
  \item Interacting with several applications
  \item Accumulating the information across the run and returning the summary to the user (i.e. the result is not visible on the last screenshot)
  \item Advanced interaction with GUI elements (drag-and-drop, selecting the parts of the text/tables)
  \item Independent resolution of potential problems (e.g. a user gives incorrect instruction)
\end{itemize} & 4 & 75 & 32 \\
\hline \\

\textbf{Gold} & \begin{itemize}
  \item Massive, open-ended tasks requiring reasoning, information accumulation, many steps
  \item Drawing with the mouse
  \item Real-time interaction with GUI (e.g. playing games that require reaction)
  \item May include multiple app interactions (like creating a spreadsheet and incorporating it into a word doc as a dynamic element via linking)
\end{itemize} & 8 & 100 & 11 \\

\bottomrule \\
\caption{Task Levels and Their Requirements\label{tab:task_levels}} \\
\end{longtable}

These levels represent the current state of the agents. The SOTA agents (at the time of writing) are expected to go through Paper and Wood with minimal issues, complete most of the Bronze tasks, show moderate performance on the Silver ones, and mostly fail on the Gold level.

To reflect the fact that these tasks vary significantly in complexity, we introduce the \textit{weighted score}, which is defined as follows.

\[
\text{Score} = 
\frac{
\sum_{\text{level} \in \{\text{Paper},\, \text{Wood},\, \text{Bronze},\, \text{Silver},\, \text{Gold}\}} 
\left( N^{\text{passed}}_{\text{level}} \cdot \text{Weight}_{\text{level}} \right)
}{
\sum_{\text{level} \in \{\text{Paper},\, \text{Wood},\, \text{Bronze},\, \text{Silver},\, \text{Gold}\}} 
\left( N_{\text{level}} \cdot \text{Weight}_{\text{level}} \right)
} \times 100\%
\]

where 

\begin{itemize}
    \item $\text{N}_{\text{level}}$ is amount of tests of the given level of complexity;
    \item $\text{N}^{\text{passed}}_{\text{level}}$ is amount of tests of the given level of complexity that an agent passed;
    \item $\text{Weight}_{\text{level}}$ is defined in Table~\ref{tab:task_levels}.
\end{itemize}

\section{SOTA Agents Scores}

We run the benchmark against 8 configurations of agents and multimodal LLMs, as described in Table~\ref{tab:agent_performance}. In all cases, the maximum number of steps is limited per level, as described in Table~\ref{tab:task_levels}.

Table~\ref{tab:agent_cost} represents the corresponding performance, cost and time agents needed to run through all tests (calculated from the moment an agent starts work, not counting the time to setup the environment).

Both Table~\ref{tab:agent_performance}~and~\ref{tab:agent_cost} show the \textbf{best} runs for each agent. Tables~\ref{tab:cua},~\ref{tab:claude-cu},~\ref{tab:claude-react},~and~\ref{tab:qwen} show statistics of multiple runs of each of the top performing agents. 

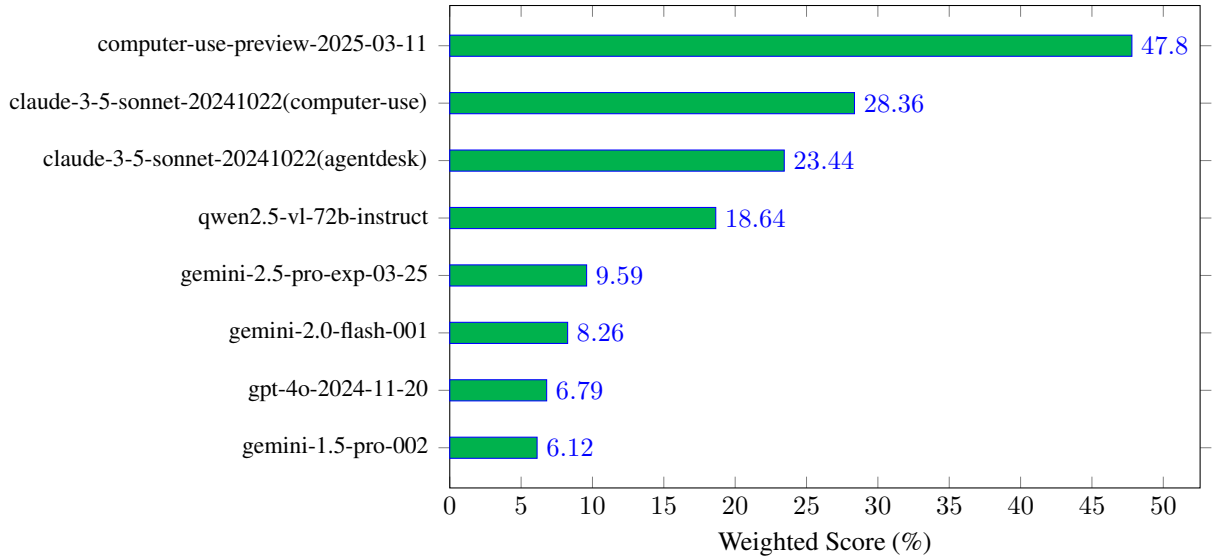
\begin{figure}
\centering

\begin{tikzpicture}
\begin{axis}[agentplottotal]
\addplot+[xbar, fill=green!70!blue] table [x=Total, y={Agent}, col sep=space] {agent_performance_data.txt};
\end{axis}
\end{tikzpicture}

\caption{Agent Performance}
\label{fig:agent_total_score}
\end{figure}

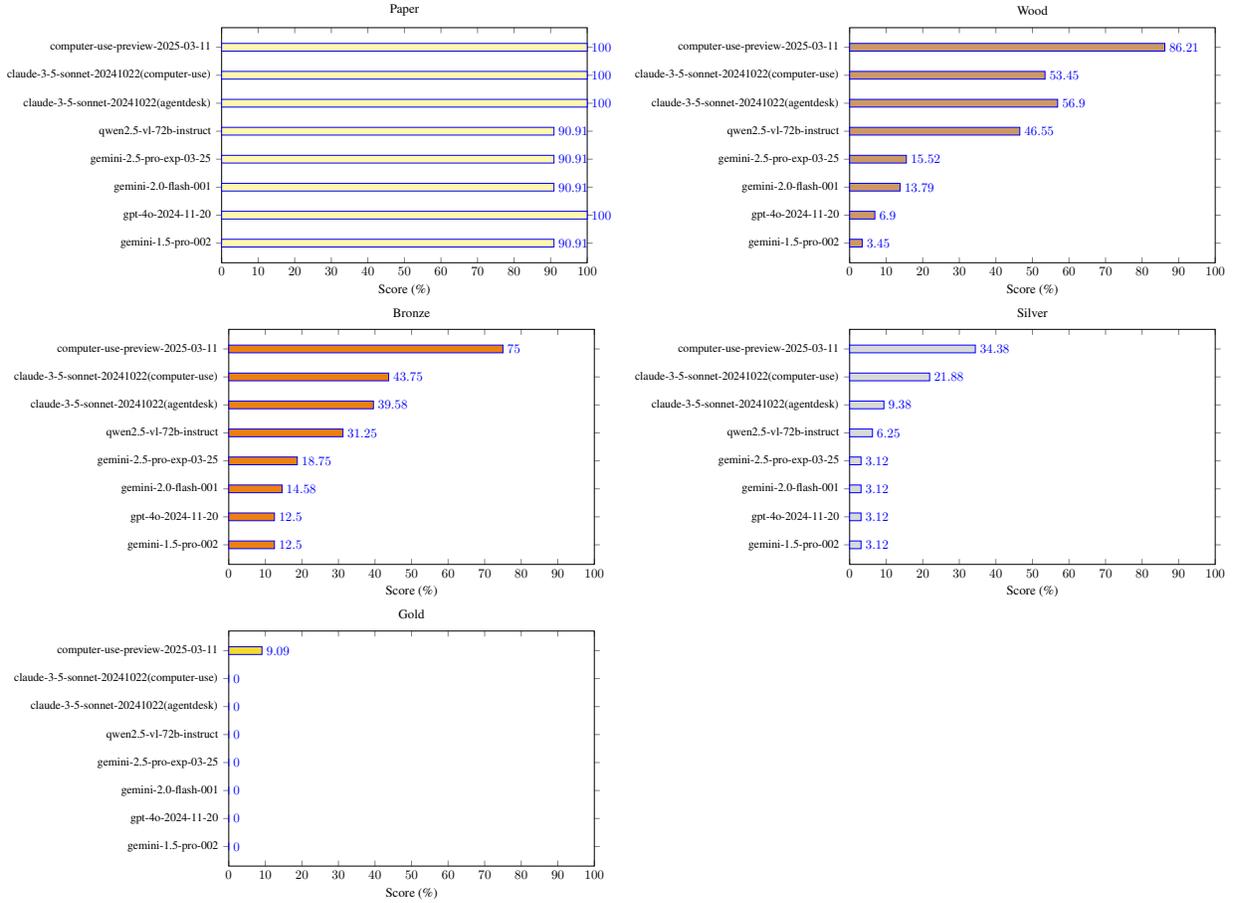
\begin{figure}
\centering

\begin{adjustbox}{max width=\textwidth}
\begin{tabular}{cc}

\begin{tikzpicture}
\begin{axis}[agentplot, title={Paper}]
\addplot+[xbar, fill=yellow!40] table [x=Paper, y={Agent}, col sep=space] {agent_performance_data.txt};
\end{axis}
\end{tikzpicture}
&
\begin{tikzpicture}
\begin{axis}[agentplot, title={Wood}]
\addplot+[xbar, fill=brown!80] table [x=Wood, y={Agent}, col sep=space] {agent_performance_data.txt};
\end{axis}
\end{tikzpicture}
\\

\begin{tikzpicture}
\begin{axis}[agentplot, title={Bronze}]
\addplot+[xbar, fill=orange!70!brown] table [x=Bronze, y={Agent}, col sep=space] {agent_performance_data.txt};
\end{axis}
\end{tikzpicture}
&
\begin{tikzpicture}
\begin{axis}[agentplot, title={Silver}]
\addplot+[xbar, fill=gray!30] table [x=Silver, y={Agent}, col sep=space] {agent_performance_data.txt};
\end{axis}
\end{tikzpicture}
\\

\begin{tikzpicture}
\begin{axis}[agentplot, title={Gold}]
\addplot+[xbar, fill=yellow!80!brown] table [x=Gold, y={Agent}, col sep=space] {agent_performance_data.txt};
\end{axis}
\end{tikzpicture}
\\

\end{tabular}
\end{adjustbox}

\caption{Agent Performance Across All Levels}
\label{fig:agent_level_score}
\end{figure}

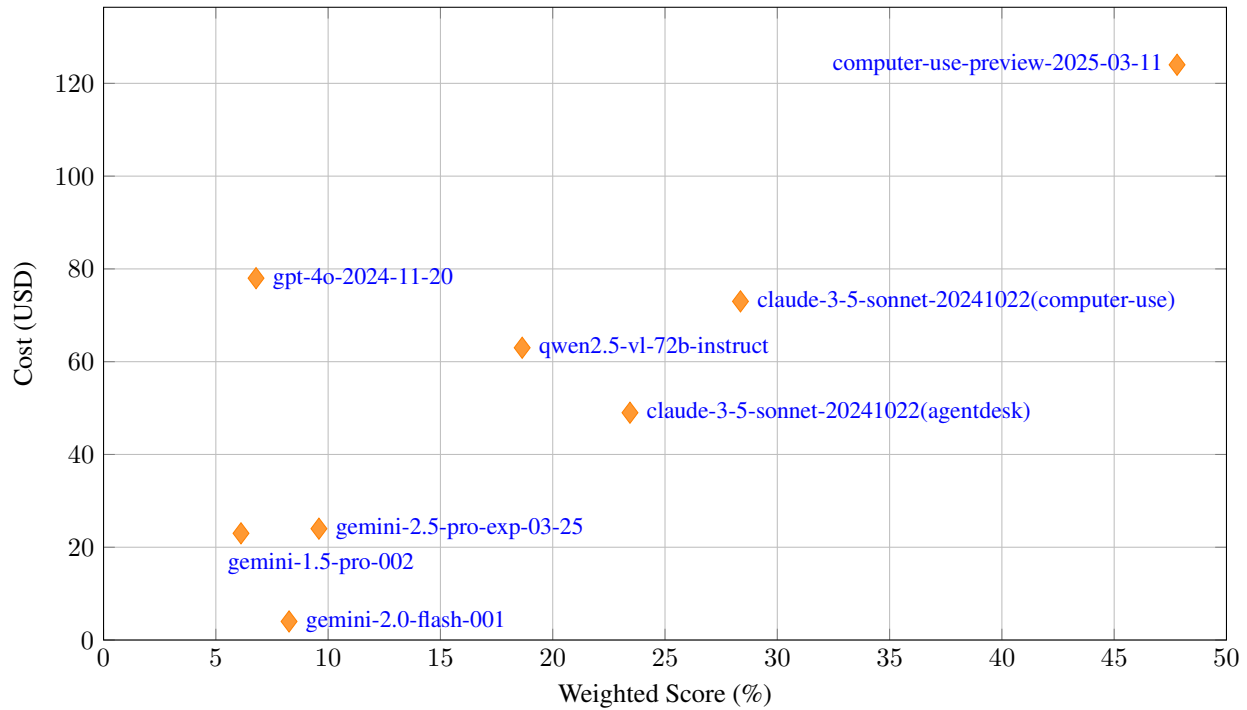
\begin{figure}

\begin{tikzpicture}
\begin{axis}[
    xlabel={Weighted Score (\%)},
    ylabel={Cost (USD)},
    width=\textwidth,        
    height=10cm,                 
    xmin=0,
    ymin=0,
    xmax=50,
    grid=both,
]

\addplot[
    scatter,
    only marks,
    mark=diamond*,
    mark size=4pt,
    draw=orange,
    fill=orange!80,
    scatter src=explicit,
    nodes near coords,
    point meta=explicit symbolic,
    nodes near coords align={horizontal},
    visualization depends on={value \thisrow{xoffset} \as \xoffset},
    visualization depends on={value \thisrow{yoffset} \as \yoffset},
    every node near coord/.append style={
        font=\small, 
        anchor=west, 
        color=blue, 
        xshift=\xoffset cm,
        yshift=\yoffset cm},
]
table[
    x=Total,
    y=Cost,
    meta=Label,
    col sep=space
] {agent_performance_data.txt};

\end{axis}
\end{tikzpicture}

\caption{Agent Performance vs Cost}
\label{fig:agent_cost}
\end{figure}

Figures~\ref{fig:agent_total_score},~\ref{fig:agent_level_score}~and~\ref{fig:agent_cost} represent the same data visually. 


\begin{longtable}{lrrrrrr}

\toprule
Agent & Paper & Wood & Bronze & Silver & Gold & Total Score \\
\midrule
\endfirsthead

\toprule
Agent (cont) & Paper & Wood & Bronze & Silver & Gold & Total Score \\
\midrule
\endhead

Computer Use Agent with \\ \textbf{computer-use-preview-2025-03-11}     & 100.00\% & 86.21\% & 75.00\% & 34.38\% & 9.09\% & \textbf{47.80\%} \\
\hline \\
Claude Computer Use with \\ \textbf{claude-3-5-sonnet-20241022}         & 100.00\% & 53.45\% & 43.75\% & 21.88\% & 0\% & \textbf{28.36\%} \\
\hline \\
AgentDesk-based ReACT with \\ \textbf{claude-3-5-sonnet-20241022}       & 90.91\%  & 56.90\% & 39.58\% &  9.38\% & 0\% & \textbf{23.44\%} \\
\hline \\
QWEN-based ReACT with \\ \textbf{qwen2.5-vl-72b-instruct}               & 90.91\%  & 46.55\% & 31.25\% & 6.25\%  & 0\% & \textbf{18.64\%} \\
\hline \\
AgentDesk-based ReACT with \\ \textbf{gemini-2.5-pro-exp-03-25}         & 90.91\%  & 15.52\% & 18.75\% & 3.12\%  & 0\% & \textbf{9.59\%} \\
\hline \\
AgentDesk-based ReACT with \\ \textbf{gemini-2.0-flash-001}             & 90.91\%  & 13.79\% & 14.58\% & 3.12\%  & 0\% & \textbf{8.26\%} \\
\hline \\
AgentDesk-based ReACT with \\ \textbf{gpt-4o-2024-11-20}                & 100.00\% & 6.90\%  & 12.50\% & 3.12\%  & 0\% & \textbf{6.79\%} \\
\hline \\
AgentDesk-based ReACT with \\ \textbf{gemini-1.5-pro-002}               & 90.91\%  & 3.45\%  & 12.50\% & 3.12\%  & 0\% & \textbf{6.12\%} \\

\bottomrule \\
\caption{Agent Performance\label{tab:agent_performance}} \\
\end{longtable}


\begin{longtable}{lrrrrr}

\toprule
Agent & Tokens In & Tokens Out & Total Score & Cost (\$) & Duration (s) \\
\midrule
\endfirsthead

\toprule
Agent (cont) & Tokens In & Tokens Out & Total Score & Cost (\$) & Duration (s) \\
\midrule
\endhead

Computer Use Agent with \\ \textbf{computer-use-preview-2025-03-11} & 40.27M & 0.25M & \textbf{47.80\%} & 124 & 55921 \\
\hline \\
Claude Computer Use with \\ \textbf{claude-3-5-sonnet-20241022}     & 23.11M & 0.24M & \textbf{28.36\%} & 73  & 26611 \\
\hline \\
AgentDesk-based ReACT with \\ \textbf{claude-3-5-sonnet-20241022}   & 14.76M & 0.39M & \textbf{23.44\%} & 49  & 29335 \\
\hline \\
QWEN-based ReACT with \\ \textbf{qwen2.5-vl-72b-instruct}           & 30.27M & 0.47M & \textbf{18.64\%} & 63  & 96500 \\
\hline \\
AgentDesk-based ReACT with \\ \textbf{gemini-2.5-pro-exp-03-25}     & 12.90M & 0.82M & \textbf{9.59\%}  & 24  & 30997 \\
\hline \\
AgentDesk-based ReACT with \\ \textbf{gemini-2.0-flash-001}         & 36.24M & 0.89M & \textbf{8.26\%}  & 4   & 45558 \\
\hline \\
AgentDesk-based ReACT with \\ \textbf{gpt-4o-2024-11-20}            & 28.82M & 0.62M & \textbf{6.79\%}  & 78  & 50337 \\
\hline \\
AgentDesk-based ReACT with \\ \textbf{gemini-1.5-pro-002}           & 15.96M & 0.41M & \textbf{6.12\%}  & 23  & 29673 \\

\bottomrule \\
\caption{Agent Performance vs Cost vs Time\label{tab:agent_cost}} \\
\end{longtable}


\begin{longtable}{lrrrrrr}

\toprule
Run & Paper & Wood & Bronze & Silver & Gold & Total Score \\
\midrule
\endfirsthead

\toprule
Run (cont) & Paper & Wood & Bronze & Silver & Gold & Total Score \\
\midrule
\endhead

\#1 & 100.00\% & 86.21\% & 75.00\% & 34.38\% & 9.09\% & \textbf{47.80\%} \\
\#2 & 100.00\% & 86.76\% & 75.00\% & 34.38\% & 0.00\% & \textbf{45.14\%} \\
\#3 &  90.00\% & 84.48\% & 70.83\% & 21.25\% & 0.00\% & \textbf{43.07\%} \\
\midrule
Mean       & 96.67\% & 85.82\% & 73.61\% & 30.00\% &  3.03\% & \textbf{45.34\%} \\
Std.\ Dev. &  5.77\% &  1.19\% &  2.41\% &  7.58\% &  5.25\% & \textbf{ 2.37\%} \\

\bottomrule \\
\caption{Computer Use Agent from OpenAI: Multiple Runs\label{tab:cua}} \\
\end{longtable}


\begin{longtable}{lrrrrrr}

\toprule
Run & Paper & Wood & Bronze & Silver & Gold & Total Score \\
\midrule
\endfirsthead

\toprule
Run (cont) & Paper & Wood & Bronze & Silver & Gold & Total Score \\
\midrule
\endhead

\#1 & 100.00\% & 53.45\% & 43.75\% & 21.88\% & 0.00\% & \textbf{28.36\%} \\
\#2 & 100.00\% & 53.45\% & 43.75\% & 12.50\% & 0.00\% & \textbf{25.17\%} \\
\#3 & 100.00\% & 55.17\% & 39.58\% & 12.50\% & 0.00\% & \textbf{24.37\%} \\
\midrule
Mean      & 100.00\% & 54.02\% & 42.36\% & 15.63\% &  0.00\% & \textbf{25.97\%} \\
Std.\ Dev.&   0.00\% &  0.99\% &  2.41\% &  5.42\% &  0.00\% & \textbf{ 2.11\%} \\

\bottomrule \\
\caption{Claude Computer Use Agent: Multiple Runs\label{tab:claude-cu}} \\
\end{longtable}


\begin{longtable}{lrrrrrr}

\toprule
Run & Paper & Wood & Bronze & Silver & Gold & Total Score \\
\midrule
\endfirsthead

\toprule
Run (cont) & Paper & Wood & Bronze & Silver & Gold & Total Score \\
\midrule
\endhead

\#1 & 100.00\% & 55.17\% & 35.42\% & 12.50\% & 0.00\% & \textbf{23.30\%} \\
\#2 & 100.00\% & 53.45\% & 35.42\% &  9.38\% & 0.00\% & \textbf{21.97\%} \\
\#3 &  90.91\% & 56.90\% & 39.58\% &  9.38\% & 0.00\% & \textbf{23.44\%} \\
\midrule
Mean      & 96.97\% & 55.17\% & 36.81\% & 10.42\% & 0.00\% & \textbf{22.90\%} \\
Std.\ Dev.&  5.25\% &  1.73\% &  2.40\% &  1.80\% & 0.00\% & \textbf{ 0.81\%} \\

\bottomrule \\
\caption{Claude Computer Use with AgentDesk action space: Multiple Runs\label{tab:claude-react}} \\
\end{longtable}


\begin{longtable}{lrrrrrr}

\toprule
Run & Paper & Wood & Bronze & Silver & Gold & Total Score \\
\midrule
\endfirsthead

\toprule
Run (cont) & Paper & Wood & Bronze & Silver & Gold & Total Score \\
\midrule
\endhead

\#1 & 90.91\%  & 46.55\% & 31.25\% & 6.25\%  & 0.00\% & \textbf{18.64\%} \\
\#2 & 90.91\%  & 48.28\% & 29.17\% & 6.25\%  & 0.00\% & \textbf{18.38\%} \\
\#3 & 90.91\%  & 48.28\% & 29.17\% & 6.25\%  & 0.00\% & \textbf{18.38\%} \\
\midrule
Mean      & 90.91\% & 47.70\% & 29.86\% & 6.25\% & 0.00\% & \textbf{18.47\%} \\
Std.\ Dev.&  0.00\% &  1.00\% &  1.20\% & 0.00\% & 0.00\% & \textbf{ 0.15\%} \\

\bottomrule \\
\caption{QWEN-based Agent: Multiple Runs\label{tab:qwen}} \\
\end{longtable}


We can share some observations and draw some conclusions.

\begin{itemize}
    \item \textbf{Computer Use Preview} from OpenAI is undoubtedly a  very spectacular model with a lot of potential for GUI navigation. At the same time, it shows the highest instability in the results among the top four agents.
    \item The agentic code in the benchmark is minimalistic and relies on the models and the sample codes from the corresponding providers (if present). This means that more elaborate agents based on this model are likely to score much higher on this benchmark, around 70-80\%. 
    \item We observed quite a few cases in which a model ran out of steps before completing the task. Relaxing the steps limits will also improve the score.
    \item At the same time, \textbf{Computer Use Preview} is the most expensive and slow model due to reasoning capabilities that are part of its decision-making. Speed in real-world scenarios is a major factor for a strong GUI-navigation agent and may be taken into account in future versions of the benchmark.
    \item All tested models demonstrate the ability to read the data in the screenshot and to return the result in the expected format.
    \item All the top models (\textbf{Computer Use Preview}, \textbf{Claude Computer Use}, and \textbf{QWEN 2.5 VL 72B}) are specifically trained for GUI navigation, and each of those models have their own action space. That is why we implemented specialized agents for each of these models. 
    \item At the same time, we can see that a strong model like \textbf{Claude Computer Use} can generalize well enough to operate in a different action space. The results in a non-native action space are slightly worse, but comparable, offering hints at the future of GUI navigation models and their potential for generalization.  
    \item \textbf{Gemini 2.0 Flash} is the cheapest model on this list; although the results are not particularly good, there is a potential to fine-tune the model and introduce reasoning at inference time to an agent. 
    \item \textbf{QWEN 2.5 VL 72B} is the only open-weight model on the list. This points to a problem where open source and open weights models have focused almost exclusively on text and multimodal models are only beginning to take focus in the open source and open weights space. There is a distinct lack of GUI-navigation-oriented multimodal models on the market today.  
    \item There is a large potential to fine-tune open source and open weight models that is lacking in proprietary models, which offers the tantalizing prospect of significantly enhancing the performance on GUI reasoning and GUI navigation tasks. For instance, much of the SOTA work on models has shifted to Reinforcement Learning (RL), kicked off by the rise of DeepSeek \cite{shao2024deepseekmathpushinglimitsmathematical} R1 \cite{deepseekai2025deepseekr1incentivizingreasoningcapability}. Currently most proprietary models are lagging in their ability to be fine-tuned with SOTA RL training methodologies such as GRPO (Group Relative Policy Optimization) and DAPO (Decoupled Clip and Dynamic sAmping Policy Optimization) \cite{yu2025dapoopensourcellmreinforcement} and REINFORCE++ \cite{hu2025reinforceefficientrlhfalgorithm}. GPT models can only be tuned with Supervised Fine Tuning (SFT) and Direct Preference Optimization (DPO). Many proprietary models cannot be fine tuned at all (many of the Claude models and the OpenAI o1-o4 line). We expect proprietary models to continue to lag in SOTA fine-tuning due to the time and cost to keep up with the sheer range of available tuning possibilities and this is likely to give open source and open weights models a significant advantage in the future on GUI navigation tasks.
    \item It is also worth noting that \textbf{QWEN 2.5 VL 72B} is the most stable model across the top four agents and demonstrates very low variation in the results.  
\end{itemize}

The main conclusion we draw is that \textbf{the best performing GUI-navigation agents are based on the models that are trained specifically for this use case. As a consequence, these models have their own action space and need custom agentic code to demonstrate their full potential.}


\section{Remarks on Automated Validation}

When we designed this benchmark, our goal was to provide enough diversity among tasks and ensure that we are not limited by the tasks that can be validated deterministically (like finding a very particular fact or creating a very particular file). The majority of tasks we face in the digital environment are not deterministic: we search for data that depends on time and location; we use various versions of tools to produce the results that comply with our requirements; we find solutions for problems that imply many possible solutions, etc.

As one might expect, nondeterministic solutions are hard to validate automatically; an obvious approach here is to use human validation. However, the issue is that human validation is not scalable: If one reruns the benchmark reasonably often while improving their agent, validating the results manually quickly becomes a tedious and time-consuming task.

To address this problem, we introduce an automated validator based on Gemini models. We chose the Gemini family because of the combination of multi-modality and context window size. This combination becomes crucial for validating the \textit{ExpectedFlowCheck}s described in Section~\ref{sec:architecture}, because to validate a result we have to send the LLM all the images from all states of the run trajectory, which can quickly saturate a small context window.

To measure the effectiveness of automated validation, we performed a human review of each of the agent runs and counted the number of tests in which the human verdict is different from the AI verdict. This number, divided by the total number of tests (160), is called the\textit{ \textbf{disagreement rate}}. 

In Table~\ref{tab:validation} you can see the disagreement rate for two models: Gemini 2.0 Flash (\textit{gemini-2.0-flash-001}), which we use as the main validation model, and Gemini 2.5 Pro Preview (\textit{gemini-2.5-pro-preview-03-25}), which we tested as an alternative.

To keep the estimation fair, we calculated the disagreement rate in eight runs from different combinations of agents and models.

It is important to note that:

\begin{itemize}
    \item Gemini 2.5 Pro Preview has noticeable rate limits at the time of writing; therefore, using it for a benchmark run is not reliable enough.
    \item Gemini 2.5 Pro Preview is slightly different from Gemini 2.0 Flash in terms of accuracy and details. One cannot drop-in replace the latter with the former; extra work with the prompts is required. 
\end{itemize}

\begin{longtable}{lrr}

\toprule
Run \# & Gemini 2.0 Flash & Gemini 2.5 Pro Preview \\
\midrule
\endfirsthead

\toprule
Run \# (cont) & Gemini 2.0 Flash & Gemini 2.5 Pro Preview \\
\midrule
\endhead

1 & 1.250\% & 1.250\% \\
2 & 1.875\% & 3.125\% \\
3 & 1.250\% & 0.000\% \\
4 & 2.500\% & 0.000\% \\
5 & 1.875\% & 0.625\% \\
6 & 0.625\% & 0.000\% \\
7 & 2.500\% & 1.250\% \\
8 & 1.250\% & 2.500\% \\
\hline \\
\textbf{Average} & \textbf{1.64\%} & \textbf{1.09\%} \\

\bottomrule \\
\caption{Validation Disagreement Rates\label{tab:validation}} \\
\end{longtable}

You can see that the average disagreement rate during our experiments was 1.64\%, which allows us to draw very reliable conclusions about the performance of agentic models without having to manually validate the runs.

The latest Gemini model shows even better results. We will keep using \textbf{Gemini 2.0 Flash} until \textbf{Gemini 2.5 Pro Preview} goes out of preview mode and can be used at scale. 

\section{Conclusions and Future Work Directions}

Our core contributions in this paper are as follows:

\begin{itemize}
    \item Presenting a new benchmark for multimodal desktop-oriented GUI-navigation AI agents.
    \item Providing a nondeterministic mechanism of automated validation with an error rate less than 2\%.
    \item Measurement of performance of several top proprietary and open-weight models. 
\end{itemize}

It is clear that this is only the beginning. There is a lot of development that can be done to improve both the benchmark and the agents.

\begin{itemize}
    \item The current SOTA is slightly below 50\%, and we may expect this benchmark to be saturated in the middle-term future. We need to create more tests, especially in the Silver and Gold categories, which would significantly increase the failure rate for today's best models.  We will leave this to version 2 of the benchmark, as these tests are challenging to create and we wanted to release a strong baseline benchmark as a foundation for later work.
    \item We also need to cover more applications and use cases. The current set of tests is limited by applications and websites that can be used without setting up accounts. More diversity is needed.
    \item Our agentic implementations represent the baseline; it is obvious that better prompting, short- and long-term memory, additional tools, and other applied AI approaches could meaningfully improve the agents' performance.
\end{itemize}

The source code of the benchmark is available at \href{https://github.com/agentsea/osuniverse}{https://github.com/agentsea/osuniverse}. You can find all the prompts and instructions there. All of the work in this paper is designed to be reproducible. If you encounter issues running the benchmark, please let us know. 

We highly encourage the community to contribute tests and implementations of agents.  

\bibliographystyle{unsrt}  
\bibliography{references}  

\end{document}